\documentclass[runningheads]{llncs}

\usepackage{eccv}

\usepackage{eccvabbrv}

\usepackage{graphicx}
\graphicspath{{alg_fig_table/}}

\usepackage{soul}
\usepackage{booktabs}
\usepackage{multirow}
\usepackage{ragged2e}

\usepackage[accsupp]{axessibility}  
\usepackage{makecell}

\usepackage{amsmath} 

\usepackage{hyperref}

\usepackage{orcidlink}

\definecolor{myredhl}{RGB}{255,180,180}

\def\titlePrefix{FluSplat}

\begin{document}

\title{FluSplat: Sparse-View 3D Editing without Test-Time Optimization} 

\titlerunning{FluSplat}

\author{Haitao Huang \and
Shin-Fang Chng \and 
Huangying Zhan \and
Qingan Yan \and
Yi Xu
}

\authorrunning{Huang et al.}

\institute{Goertek Alpha Labs}

\maketitle
\begin{abstract}
Recent advances in text-guided image editing and 3D Gaussian Splatting (3DGS) have enabled high-quality 3D scene manipulation. 
However, existing pipelines rely on iterative edit-and-fit optimization at test time, alternating between 2D diffusion editing and 3D reconstruction. 
This process is computationally expensive, scene-specific, and prone to cross-view inconsistencies.

We propose a feed-forward framework for cross-view consistent 3D scene editing from sparse views. 
Instead of enforcing consistency through iterative 3D refinement, we introduce a cross-view regularization scheme in the image domain during training. 
By jointly supervising multi-view edits with geometric alignment constraints, our model produces view-consistent results without per-scene optimization at inference. 
The edited views are then lifted into 3D via a feedforward 3DGS model, yielding a coherent 3DGS representation in a single forward pass.

Experiments demonstrate competitive editing fidelity and substantially improved cross-view consistency compared to optimization-based methods, while reducing inference time by orders of magnitude.

    \keywords{Text-Guided 3D Scene Editing \and Sparse-View Novel View Synthesis \and Cross-View Consistenct Editing}
\end{abstract}

\section{Introduction}
\label{sec:intro}

The ability to 	\textit{efficiently} edit 3D scenes in a semantically controlled and visually coherent manner is fundamental to immersive media, extended reality, and robotics simulation. 
Unlike 2D image editing, 3D scene editing must preserve consistency across viewpoints: even small cross view discepancies inevitably amplify under novel-view rendering, breaking spatial realism and immersion. 
Multi-view coherence is therefore not optional but intrinsic to 3D editing.

Recent advances in diffusion and rectified-flow models have enabled high-fidelity text-guided image editing, while 3D scene representations—from Neural Radiance Fields (NeRF)~\cite{mildenhall2021nerf, mvsnerf, vicanerf, barf, nope-nerf, garf,wang2021nerf} to 3D Gaussian Splatting (3DGS)~\cite{kerbl20233dgs}—support scalable reconstruction and real-time novel-view synthesis. Bridging these advances, however, remains challenging. Modern 2D editors operate independently on images without explicit 3D awareness; when applied across views and lifted to 3D, they introduce inconsistencies that amplify under novel viewpoints. Consequently, existing text-guided 3D editing methods incur substantial overhead to enforce coherence.

Current 3DGS editing approaches mainly differ in \textit{where} coherence is enforced. Some rely on iterative edit-and-fit loops at test time~\cite{igs2gs,haque2023instruct}, while others introduce multi-view coupling within the editor or conditioning~\cite{chen2024dge, chen2024gaussianeditor, lee2025editsplat, chen2025vip3dedit}, often requiring per-scene tuning. Supervised fine-tuning~\cite{zhao2025tinker} reduces test-time optimization but depends on curated training data. Despite these variations, prior methods still incur scene-specific optimization or radiance field refinement, limiting efficiency in sparse-view or real-time settings.

We revisit sparse-view 3D editing from a different perspective. Rather than optimizing geometry iteratively or relying on curated supervision, we focus on resolving cross-view inconsistencies directly in the image domain. Once edits are made consistent across views, geometry can be reconstructed reliably in a feedforward manner.

Motivated by this insight, we propose \titlePrefix{}, a fully \textit{feedforward sparse-view 3D editing} framework that operates from only two input views and a text instruction.
Our key idea is to decouple editing from reconstruction: we first make the edits cross-view consistent in the image domain, then lift the aligned results into 3D. 
Concretely, \titlePrefix{} consists of two stages:
(1) \textit{self-supervised} cross-view consistent image editing via feature consistency regularization, and 
(2) \textit{pose-free} lifting into a canonical 3D Gaussian representation using a feedforward generalizable 3DGS. 
Crucially, both editing and reconstruction are performed in a single forward pass, eliminating per-scene optimization~\cite{igs2gs} and manual hyperparameter tuning~\cite{chen2024dge} at inference time.

To enforce cross-view coherence, we fine-tune a rectified-flow editing backbone with complementary \textit{global and local feature alignment} objectives.
The global diffusion feature loss aligns high-level structural representations across views, while the proposed local editing feature loss constrains region-specific consistency for localized edits.
Importantly, this fine-tuning is fully self-supervised, relying solely on cross-view feature agreement.
We adopt Flux~\cite{labs2025flux} for its strong semantic editing capability.
Once 2D alignment is achieved, the edited views are lifted into 3D using a pose-free NoPoSplat-style feedforward Gaussian reconstruction network, producing consistent and editable 3D scenes in one pass.

Extensive experiments on object-centric (DTU~\cite{aanaes2016dtu},IN2N~\cite{haque2023instruct}) and large-scale scene datasets (RE10K~\cite{zhou2018stereo}) demonstrate that \titlePrefix{} achieves superior 3D semantic consistency under novel viewpoints while reducing editing time by an order of magnitude compared to prior pipelines. 
Despite being trained on a single dataset, our method generalizes robustly across diverse scene types and camera distributions, highlighting the effectiveness and robustness of cross-view regularized editing as a transferable inductive bias.

In summary, our contributions are:
\vspace{-0.5em}
\begin{itemize}
    \item We introduce the \textit{first fully feedforward sparse-view 3D editing framework} that eliminates per-scene optimization.
    \item We propose a \textit{self-supervised} cross-view diffusion regularization, including global and local feature alignment, to enforce multi-view editing consistency.
    \item We demonstrate that resolving cross-view inconsistencies in the image domain enables direct 3D lifting, achieving superior 3D semantic fidelity and significantly improved efficiency.
\end{itemize}

\section{Related Work}
\label{sec:rel_work}

\paragraph{\textbf{2D Editing}}
Text-guided diffusion models such as Stable Diffusion~\cite{stablediffusion}, 
Imagen~\cite{Imggen}, and DALL-E 2~\cite{dalle} dominate high-fidelity image generation and editing, 
aiming to apply semantic changes while preserving structure and identity. 
SDEdit~\cite{SDEdit} formulates editing as controlled denoising, 
trading edit strength for fidelity, a principle underlying modern \textit{img2img} pipelines 
but highlighting the tension between stronger edits and structural drift.

Editing-by-inversion methods~\cite{Nulltextinversion, masactrl} 
map images into latent space before re-denoising with modified prompts; 
null-text inversion~\cite{Nulltextinversion} improves reconstruction fidelity. 
Inversion-free approaches such as InstructPix2Pix~\cite{instructpix2pix} 
and follow-ups~\cite{boesel2024improving} enable instruction-following editing via synthetic supervision. 
Subsequent editors—including EmuEdit~\cite{sheynin2024emuedit}, 
OmniGen~\cite{xiao2025omnigen}, 
HiDream-I1~\cite{cai2025hidream}, 
and ICEdit~\cite{zhang2025enabling}—further refine this paradigm through improved data and architectures.

Recent flow and rectified-flow models~\cite{kulikov2025flowedit, rout2024semantic, wang2024rfsolver} 
extend editing to continuous-time formulations. 
In-context editors such as FLUX.1 Kontext~\cite{labs2025flux} 
unify generation and editing via joint image-text conditioning. 
However, existing 2D methods lack explicit \textit{multi-view} consistency, 
which is essential for 3D editing. 
Building on FLUX.1 Kontext, we introduce consistency-aware regularization 
to produce view-consistent edits.

\paragraph{\textbf{3D Editing}}

Text-guided 3D editing integrates 2D diffusion editors with 3D representations such as NeRF~\cite{mildenhall2021nerf} and 3D Gaussian Splatting (3DGS)~\cite{kerbl20233dgs}. Early text-to-3D methods directly optimized 3D representations under diffusion priors. DreamFusion~\cite{poole2022dreamfusion}, for example, applies score distillation to iteratively update a radiance field so that its renderings match a text-conditioned diffusion model.
A widely adopted alternative follows a \emph{``2D edit $\rightarrow$ 3D reconcile''} pipeline: render views from a 3D scene, edit them with an image-conditioned diffusion model, and optimize the scene to absorb the edits. InstructNeRF~\cite{haque2023instruct}, InstructGS2GS~\cite{igs2gs}, and related variants~\cite{mirzaei2024watch, chen2024gaussianeditor} alternate between rendering, instruction-guided editing, and scene optimization. Although flexible, this iterative process is computationally expensive and prone to cross-view inconsistencies.

To enhance multi-view coherence, DGE~\cite{chen2024dge}, EditSplat~\cite{lee2025editsplat}, and GaussCtrl~\cite{wu2024gaussctrl} explicitly couple edits across viewpoints. However, these approaches still require scene-specific tuning, with hyperparameters tied to edit strength, inversion schedules, masking thresholds, or reference views. TINKER~\cite{zhao2025tinker} generates dense multi-view consistent edits via DiT-based diffusion and an any-view-to-video synthesizer, followed by per-scene 3DGS reconstruction, but relies on curated paired data supervision and still requires scene-level reconstruction at inference. In contrast, our method enables fully feedforward \emph{sparse-view} editing while maintaining multi-view consistency.

\paragraph{\textbf{Generalizable Sparse View Novel-view synthesis}}
NeRF~\cite{mildenhall2021nerf, murf, mvsnerf, tensorf, kplanes, barf, garf,nope-nerf,wang2021nerf} and 3D Gaussian Splatting (3DGS)~\cite{kerbl20233dgs, 2DGS} have greatly advanced 3D reconstruction and novel-view synthesis, but they typically require dense posed images and per-scene optimization. Recent work instead focuses on \textit{generalizable} sparse-view novel-view synthesis via feedforward inference~\cite{charatan2024pixelsplat, chen2024mvsplat, jiang2025anysplat, ye2024noposplat, splatt3r,depthsplat}. A common approach predicts geometry from multi-view cues using a geometry estimator~\cite{xu2023unifying, wang2024dust3r}, followed by Gaussian decoding. PixelSplat~\cite{charatan2024pixelsplat} employs epipolar cross-attention to estimate per-pixel depth distributions and sample Gaussian centers, while MVSplat~\cite{chen2024mvsplat} adopts cost-volume fusion with direct depth regression; both assume known poses. In contrast, pose-free methods such as NoPoSplat~\cite{ye2024noposplat} eliminate pose prerequisites by leveraging large-scale transformer priors (e.g., Dust3R-style correspondence reasoning~\cite{wang2024dust3r,mast3r}), achieving competitive sparse-view synthesis without explicit geometry or pose input. Our method builds upon the NoPoSplat architecture.

\section{Method}
\label{sec:method}

We introduce \titlePrefix{}, a feedforward sparse-view 3D editing framework that enables direct text-guided manipulation of 3D scenes from only two sparse views. 
Given a pair of sparse RGB images and a textual editing instruction, our method produces the edited 3D Gaussian Splatting representation~\cite{kerbl20233dgs} in a single forward pass, eliminating the need for iterative score distillation~\cite{poole2022dreamfusion}, per-scene optimization~\cite{haque2023instruct, igs2gs, chen2024dge}, or geometry refinement~\cite{haque2023instruct, igs2gs}.

\begin{figure}[t!]
    \centering
    \includegraphics[width=\linewidth]{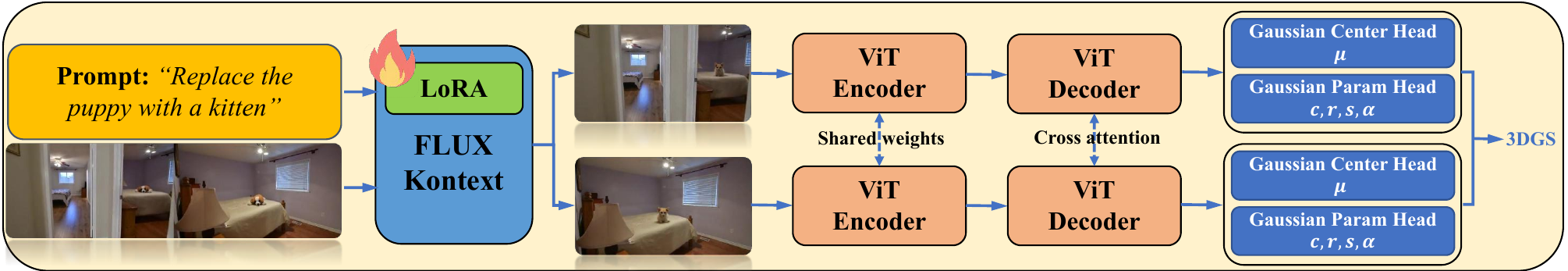}
    \caption{\textbf{\titlePrefix{} Pipeline.}
 Given two sparse-view input images and a textual editing instruction, we first apply a FLUX model\cite{labs2025flux} finetuned via LORA to generate cross-view consistent edited images. The edited images are then processed by a transformer-based sparse-view reconstruction network~\cite{ye2024noposplat} to obtain the edited 3DGS scene representation. The overall instruction-conditioned image-to-3D editing process is fully feedforward and takes about \textit{20 seconds} to complete.}
    \label{fig:pipeline}
    \vspace{-10pt}
\end{figure}

As illustrated in Fig.~\ref{fig:pipeline}, the pipeline consists of two sequential components:
(1) cross-view consistent image editing, which ensures geometrically coherent edits across sparse viewpoints, and 
(2) feedforward 3D Gaussian reconstruction, which lifts the edited views into a canonical 3D Gaussian representation via a transformer-based network.

\vspace{-10pt}
\subsection{Feedforward Sparse-View 3D Editing}
\label{sec:framework}

We consider a sparse-view scenario with two input RGB images
and their camera intrinsics, but without known camera extrinsics.
Given a textual editing prompt $c$,
the input is
$
\{ I_i, k_i, c \}_{i=0,1},
$
where $I_i \in \mathbb{R}^{H \times W \times 3}$ denotes the $i$-th image
and $k_i$ its intrinsic parameters. Our goal is to directly predict an edited 3DGS scene from these sparse inputs. 

We formulate sparse-view 3D editing as a direct conditional mapping:
\begin{equation}
f_{\theta} :
\{I_i, k_i, c\}_{i=0,1}
\rightarrow
\left\{ (\boldsymbol{\mu}_g, \alpha_g, \boldsymbol{r}_g, \boldsymbol{s}_g, \boldsymbol{c}_g) \right\}_{g=1}^{G},
\end{equation}
where the output consists of $G$ Gaussian primitives defined in canonical 3D space.
Each Gaussian $g$ is parameterized by its center $\boldsymbol{\mu}_g \in \mathbb{R}^3$, 
opacity $\alpha_g \in \mathbb{R}^{+}$,
rotation quaternion $\boldsymbol{r}_g \in \mathbb{R}^4$,
anisotropic scale $\boldsymbol{s}_g \in \mathbb{R}^3$,
and spherical harmonic coefficients $\boldsymbol{c}_g \in \mathbb{R}^d$.

Unlike optimization-based 3D editing approaches that couple geometry refinement with diffusion supervision through iterative score distillation~\cite{shapeditor,li2023dreamedit}, 
our formulation decouples editing and reconstruction and enables fully feedforward inference.
This design makes sparse-view 3D editing efficient, stable, and generalizable across scenes.

\subsection{Cross-View Consistent Image Editing}
\label{sec:editing}
Editing each sparse view independently often leads to geometric inconsistencies,
which propagate to unstable 3D reconstruction.
To address this issue, we fine-tune a pretrained image editor with explicit
cross-view consistency constraints, while preserving its original rectified-flow
generative ability.

\subsubsection{Rectified-Flow Fine-Tuning}

\begin{figure}[tbp]
    \centering
    \includegraphics[width=\linewidth]{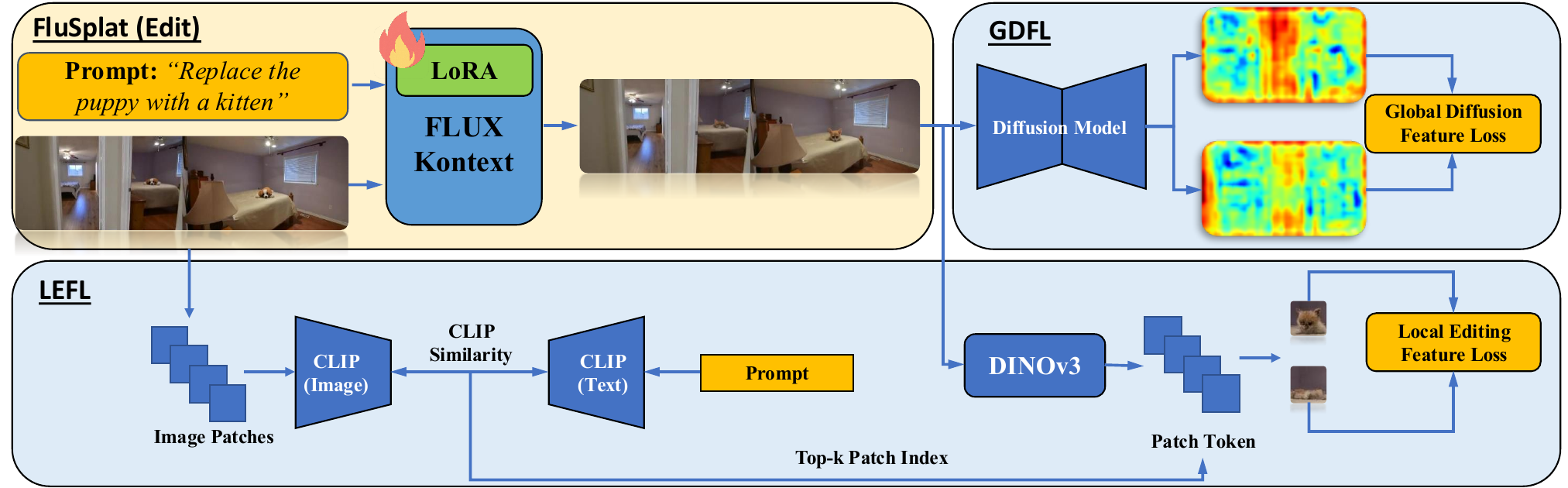}
\caption{\textbf{Cross-view consistent FLUX fine-tuning.}
A LoRA-adapted FLUX model edits two sparse views while enforcing cross-view coherence. Global Diffusion Feature Loss (GDFL) aligns intermediate diffusion features for global consistency, and Local Editing Feature Loss (LEFL) aligns CLIP-localized regions with DINOv3 features for region-level alignment. Together, these regularizations stabilize single-step editing.}
    \label{fig:finetune}
\end{figure}

We build on FLUX Kontext~\cite{labs2025flux}, a pretrained rectified-flow model that learns, 
a time-dependent velocity field that transports Gaussian noise 
to data latents via an ordinary differential equation (ODE). 
Given a latent variable $z_t$ at time $t \in [0,1]$, 
the model predicts a conditional velocity field
$v_\theta(z_t,t,c)$ under editing prompt $c$.

Importantly, we neither train the rectified-flow model from scratch 
nor rely on paired edited supervision~\cite{zhao2025tinker}.
Instead, we preserve the pretrained generative dynamics 
and fine-tune the model in a self-supervised manner using cross-view consistency regularization.
We generate edited images via single-step rectified-flow sampling, compute cross-view losses in feature space, and backpropagate their gradients through the single-step ODE update. This encourages structural alignment across views while retaining the original text-guided editing capability.


\paragraph{LoRA-Based Adaptation.}
To prevent catastrophic drift of the pretrained generative prior,
we adopt parameter-efficient Low-Rank Adaptation (LoRA)~\cite{hu2022lora}. We insert rank $r=4$ LoRA module into later transformer blocks
of both the double-stream and single-stream branches.
During training, only LoRA parameters are updated,
while the backbone weights remain frozen.
This localized low-rank adaptation provides sufficient capacity
to enforce cross-view alignment without degrading semantic fidelity in deeper layers.
A detailed architecture of the FLUX fine-tuning setup is provided in the supplementary material.

\paragraph{Single-Step Inference.}
During inference, we perturb the source latent with Gaussian noise $\epsilon \sim \mathcal{N}(0,I)$ to obtain $z_1$, and integrate the learned ODE from $t=1$ to $t=0$ via a single-step Euler update with $\Delta t = 1$:
\begin{equation}
\frac{d z_t}{dt} = v_\theta(z_t,t,c), \quad
z_0 = z_1 - v_\theta(z_1,1,c).
\end{equation}
We decode the resulting latent $z_0$ with the VAE decoder $g$ to obtain the edited image. 
Compared to multi-step diffusion sampling, this single-step rectified-flow formulation directly produces a near-clean latent representation, enabling stable decoding into the image domain for explicit loss computation and supervision.

\subsubsection{Global Diffusion Feature Loss (GDFL)}

Naïvely concatenating two views and feeding them into an image editing model
does not explicitly enforce geometric consistency, and often results in
cross-view structural drift under sparse-view settings. To enforce global cross-view alignment, we regularize intermediate diffusion features.

Given the edited views $I_0'$ and $I_1'$, we concatenate them horizontally as
\begin{equation}
I^{cat} = \mathrm{concat}(I_0', I_1') \in \mathbb{R}^{H\times2W\times3}.
\end{equation}
We then apply forward diffusion at timestep $t$ to obtain a noised input $I_t^{cat}$,
which is processed by the diffusion U-Net $f_\theta$. From a chosen layer $l$, we extract an intermediate feature map $F$.

Since the input is concatenated along the width dimension,
each feature map can be partitioned into two halves, 
$F_0$ and $F_1$, corresponding to the two views.
We define the global diffusion feature loss as
\begin{equation}
\mathcal{L}_{\text{GD}}
=
\left\| 
F_0 - F_1
\right\|_2^2.
\end{equation}
Diffusion features at moderate noise levels encode
high-level semantics while remaining sensitive to spatial layout.
Aligning these representations therefore encourages the two edited views to share
consistent global semantics and geometry.

Following Tang \textit{et al.}~\cite{tang2023emergent}, we extract feature maps from selected timesteps and network depths
to balance semantic alignment and geometric consistency:
features from earlier layers at higher noise levels capture global semantics,
whereas features from deeper layers at lower noise levels encode finer geometric details~\cite{tang2023emergent}. 

\subsubsection{Local Editing Feature Loss (LEFL)}
\label{sec:method_LEFL}

While global alignment enforces overall semantic consistency,
localized edits introduce a more challenging scenario:
the edited region must remain geometrically consistent across views
without affecting unrelated background content.
Relying solely on segmentation masks does not explicitly enforce
cross-view feature alignment within the edited region.
To address this, we introduce a Local Editing Feature Loss (LEFL),
which constrains region-level representations across views.

Our approach consists of two stages:
(i) region localization guided by text,
and (ii) cross-view local feature alignment.

\paragraph{Text-Guided Region Localization.}
Given the two pre-edit input images
$I_l, I_r \in \mathbb{R}^{H \times W \times 3}$
and the editing description $c$,
we localize the target editing region using CLIP similarity~\cite{radford2021clip}. Here, $I_l$ and  $I_r$ denote the left and right views obtained by splitting the unedited concatenated input $I^{cat} = \mathrm(I_0, I_1)$ along the width dimension.
Each image is partitioned into non-overlapping patches of size $p \times p$, yielding a grid of $\frac{H}{p} \times \frac{W}{p}$ patches.
Let $p_{ij}^l$ and $p_{ij}^r$ denote the patches from the left and right views, respectively.
We encode the text prompt $c$ as
\begin{equation}
f_T = \mathrm{TextEncoder}_{\text{CLIP}}(c),
\end{equation}
and extract the patch-level image features as
\begin{equation}
f_{ij}^l = \mathrm{ImageEncoder}_{\text{CLIP}}(p_{ij}^l),
\quad
f_{ij}^r = \mathrm{ImageEncoder}_{\text{CLIP}}(p_{ij}^r).
\end{equation}
We then compute cosine similarity between each patch feature and the text feature
to identify the most relevant patch in each view:
\begin{equation}
(i^l, j^l) =
\arg\max_{i,j}
\cos(f_{ij}^l, f_T),
\quad
(i^r, j^r) =
\arg\max_{i,j}
\cos(f_{ij}^r, f_T).
\end{equation}
These indices localize the corresponding editing regions
in both views.

\paragraph{Cross-View Local Feature Alignment.}

Using the localized regions, we extract dense features from the edited images
$I_l'$ and $I_r'$ using DINOv3~\cite{simeoni2025dinov3},
denoted as $f_{\text{DINO}}(I)$.
The local editing loss is defined as:
\begin{equation}
\mathcal{L}_{\text{LE}}
=
\left\|
f_{\text{DINO}}(I_l')[i^l, j^l]
-
f_{\text{DINO}}(I_r')[i^r, j^r]
\right\|_2^2.
\end{equation}
By enforcing alignment in a high-level self-supervised feature space,
LEFL explicitly constrains the edited regions to remain
semantically and geometrically consistent across viewpoints.

\subsubsection{Overall Editing Objective}

The final fine-tuning objective combines the proposed global and local consistency losses:

\begin{equation}
\mathcal{L}_{\text{edit}}
=
\lambda_{\text{GD}} \mathcal{L}_{\text{GD}}
+
\lambda_{\text{LE}} \mathcal{L}_{\text{LE}}.
\end{equation}

Here, 
$\mathcal{L}_{\text{GD}}$ enforces global cross-view alignment
in diffusion feature space,
and $\mathcal{L}_{\text{LE}}$ constrains region-level consistency
for localized edits.
The weighting coefficients $\lambda_{\text{GD}}$ and $\lambda_{\text{LE}}$
balance alignment constraints.

Our proposed unified objective enables stable fine-tuning without paired edited
supervision~\cite{zhao2025tinker}, while explicitly enforcing cross-view geometric coherence
under sparse-view conditions.
Detailed implementation details are provided in the supplementary material.

\subsection{Feedforward Gaussian Reconstruction}

Given cross-view consistent edited images, we reconstruct
the edited 3D scene using a feedforward Gaussian reconstruction
network based on NoPoSplat~\cite{ye2024noposplat}.
The network consists of a shared Vision Transformer (ViT) encoder,
a cross-view decoder, and Gaussian parameter prediction heads that 
directly regress 3D Gaussian primitives
from sparse-view image observations.

Concretely, each edited RGB image is partitioned into non-overlapping patches
and embedded into visual tokens. We project the camera intrinsics into learnable embeddings and concatenate them
with the iamge tokens.
The encoder processes each view independently using shared weights,
and the decoder then fuses information across views via multi-head cross-attention to enable cross-view feature interaction.

Finally, we predict Gaussian parameters from the decoded features using dedicated regression heads,
including Gaussian centers, opacity, rotation,
anisotropic scale, and spherical harmonic coefficients.
We refer readers to NoPoSplat~\cite{ye2024noposplat} for architectural details.
In our framework, the reconstruction module serves as a feedforward lifting mechanism, mapping cross-view consistent edited images into a canonical 3D Gaussian representation.

\section{Experiments and Results}
\label{sec:exp_results}

\subsection{Experimental Setup}

We evaluate \titlePrefix{} on both object-centric and large-scale scene datasets. 
IN2N~\cite{haque2023instruct} is used for 2D editing evaluation on scenes. 
DTU~\cite{aanaes2016dtu} serves for object-level 3D editing and provides calibrated multi-view images. 
RealEstate10K~\cite{zhou2018stereo} (RE10K) is used for training and for large-scale scene-level editing evaluation. 
For all datasets, we sample two sparse input views and generate edited 3D Gaussian scenes in a single forward pass.

\subsubsection{Implementation Details}
FLUX~\cite{labs2025flux} is fine-tuned on RE10K using cross-view consistency losses, see \cref{sec:editing}.
The feedforward reconstruction module adopts the original NoPoSplat~\cite{ye2024noposplat} architecture without modification.
Training is conducted on Nvidia-H200, and inference time is measured on a single Nvidia-H200.
Detailed training configurations are provided in the supplementary material.




\subsection{Evaluation Metrics}
\label{sec:metrics}

We evaluate \titlePrefix{} from three key aspects aligned with our objectives:
(i) semantic consistency with respect to the text instruction,
(ii) cross-view editing coherence, and
(iii) runtime efficiency.
All metrics are averaged over all test scenes and editing instructions.

\subsubsection{Semantic Consistency}

Our primary objective is to generate a semantically correct
edited 3D scene whose editing effects remain consistent
both on the input views and under novel viewpoints.
We therefore evaluate semantic alignment in a unified CLIP-based framework.

Let $\mathcal{V}$ denote a set of evaluated images.
For 2D evaluation, $\mathcal{V}$ corresponds to the edited input views.
For 3D evaluation, $\mathcal{V}$ consists of rendered novel views
$\{R_v\}_{v=1}^{K}$ from the reconstructed 3D Gaussian scene.

\paragraph{CLIP Similarity.}

We measure semantic alignment with the target prompt $T_1$ as:
\begin{equation}
\mathrm{CLIP}_{sim}(\mathcal{V})
=
\frac{1}{|\mathcal{V}|}
\sum_{I \in \mathcal{V}}
\cos\big(E_t(T_1), E_i(I)\big),
\end{equation}
where $E_t(\cdot)$ and $E_i(\cdot)$ denote the CLIP text and image encoders~\cite{radford2021clip}.
Higher values indicate stronger alignment between the edited results and the text prompt.

\paragraph{CLIP Directional Similarity.}

When original images $\{I^{orig}\}$ are available,
we additionally measure editing direction consistency:
\begin{equation}
\mathbf{v}_t = E_t(T_1) - E_t(T_0),
\quad
\mathbf{v}_i(I) = E_i(I) - E_i(I^{orig}).
\end{equation}
\begin{equation}
\mathrm{CLIP}_{dir}(\mathcal{V})
=
\frac{1}{|\mathcal{V}|}
\sum_{I \in \mathcal{V}}
\cos(\mathbf{v}_t, \mathbf{v}_i(I)).
\end{equation}
This unified formulation allows direct comparison
between semantic consistency at the 2D editing stage
and after 3D lifting under novel viewpoints.


\subsubsection{Cross-View 2D Consistency}

We quantify cross-view editing coherence using the same 
region-level DINO feature alignment defined in LEFL (Sec.~\ref{sec:method_LEFL}),
but treat it purely as an evaluation metric.

Specifically, we compute the $\ell_2$ distance between the 
localized DINO features of the two edited views.
Lower values indicate stronger cross-view semantic and geometric consistency.

\subsubsection{Runtime Efficiency}

We report the average editing time per scene,
including both image editing and feedforward 3D reconstruction.

\subsection{Sparse-View 3D Editing}
\label{sec:3d_editing}

\begin{figure}[!h]
    \centering
    \includegraphics[width=\linewidth]{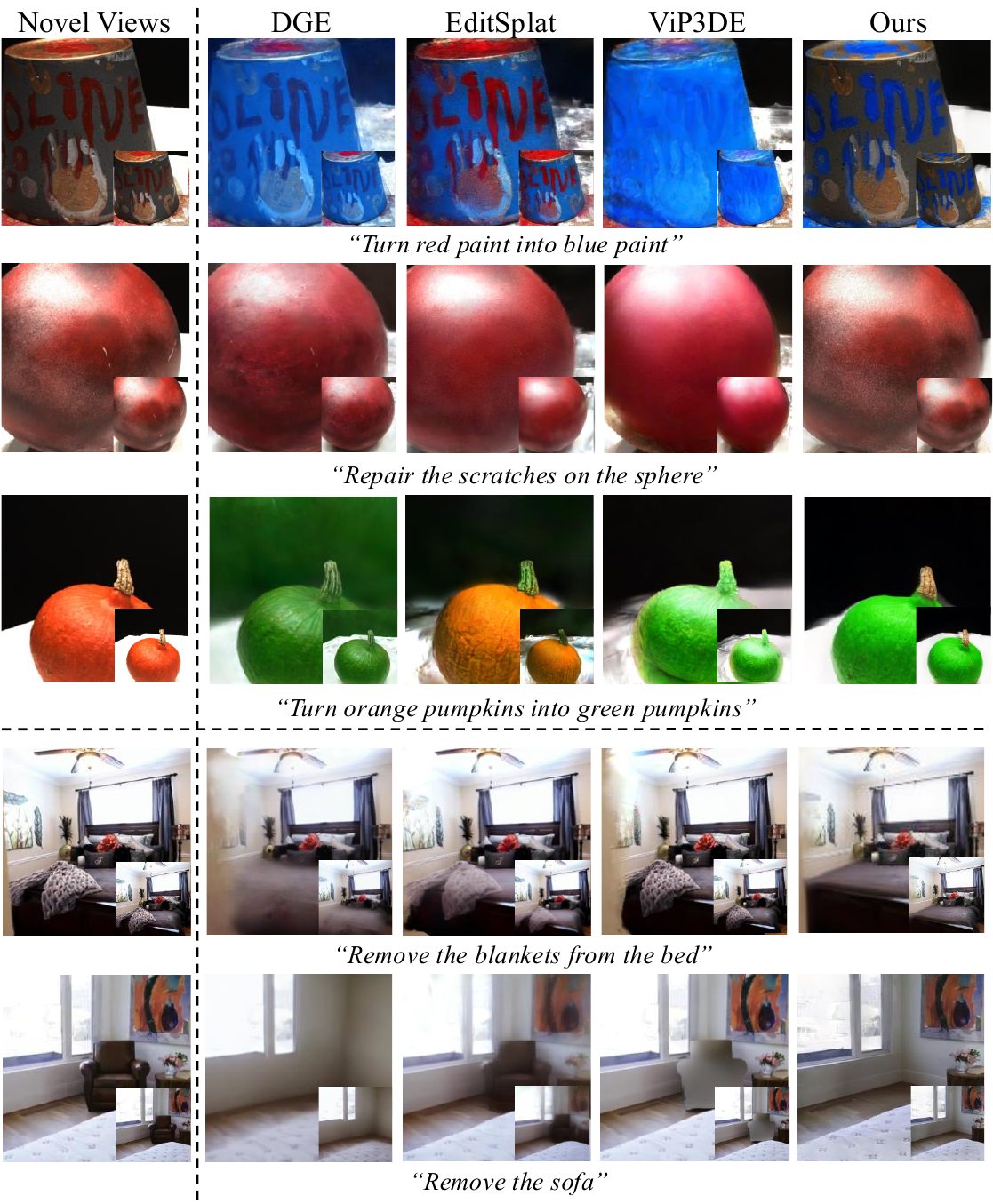}
    \caption{\textbf{Qualitative comparison on DTU~\cite{aanaes2016dtu} and RE10K~\cite{zhou2018stereo}.}
We compare \titlePrefix{} with DGE~\cite{chen2024dge}, EditSplat~\cite{lee2025editsplat}, and ViP3DE\cite{chen2025vip3dedit}
under diverse text-guided editing instructions, on DTU(top 3 rows, \emph{zero-shot}) and RE10K(bottom 2 rows, \emph{in-domain}).
For each method, the large image and its inset correspond to \emph{two distinct novel views} of the edited 3D scene.
Our method produces more consistent color transformations,
better structural preservation, fewer cross-view artifacts, and more consistent object removal than prior approaches.
In contrast, prior methods exhibit color bleeding, or incomplete edits.
}
    \label{fig:dtu_re10k}
\end{figure}


We evaluate \titlePrefix{} on text-guided sparse-view 3D editing
against representative state-of-the-art methods,
including DGE~\cite{chen2024dge}, EditSplat~\cite{lee2025editsplat}, and ViP3DE~\cite{chen2025vip3dedit}.
Unlike prior multi-stage pipelines that rely on geometric alignment
or per-scene 3D fitting, our method performs cross-view editing
and 3D reconstruction fully feedforward, without test-time optimization.

\paragraph{Evaluation Protocol.}
We conduct experiments on eight scenes from DTU and RE10K,
using two editing instructions per scene (16 scene–prompt pairs).
For each scene, our method takes two sparse input views
to produce the edited 3D Gaussian representation.
To measure semantic consistency after reconstruction,
we render $K=4$ novel views per scene,
resize them to $256 \times 256$,
and compute CLIP-based metrics averaged across views,
instructions, and scenes.
For optimization-based baselines, we follow their official implementations
and recommended settings.

\paragraph{Results.}
\begin{table}[t!]
\centering
\caption{\textbf{3D Semantic Consistency and Efficiency Comparison on RE10K and DTU.}
We report CLIP similarity and directional similarity computed on rendered novel views.
Our method consistently achieves superior 3D semantic alignment while significantly reducing editing time.}
\label{tab:3d_results}
\begin{tabular}{lcc|cc|c}
\toprule
\multirow{2}{*}{Method}
& \multicolumn{2}{c|}{\textbf{RE10K}}
& \multicolumn{2}{c|}{\textbf{DTU}}
& \multirow{2}{*}{\textbf{Time} $\downarrow$} \\
\cmidrule(lr){2-3}
\cmidrule(lr){4-5}
& CLIP$_{sim}^{3D}$ $\uparrow$
& CLIP$_{dir}^{3D}$ $\uparrow$
& CLIP$_{sim}^{3D}$ $\uparrow$
& CLIP$_{dir}^{3D}$ $\uparrow$ \\
\midrule
DGE~\cite{chen2024dge} 
& 0.195 & 0.037
& 0.211 & 0.077
& $\sim$ 4 min \\

EditSplat~\cite{lee2025editsplat}  
& 0.184 & 0.029
& 0.224 & 0.085
& $\sim$ 6 min \\

ViP3DE~\cite{chen2025vip3dedit}  
& 0.196 & 0.038
& 0.230 & 0.041
& $\sim$ 9 min \\

\midrule
Ours       
& \textbf{0.218} & \textbf{0.051}
& \textbf{0.241} & \textbf{0.089}
& \textbf{$\sim$ 20 s} \\
\bottomrule
\end{tabular}
\end{table}

As shown in Table~\ref{tab:3d_results} and Fig.~\ref{fig:dtu_re10k},
\titlePrefix{} achieves superior 3D semantic consistency
while reducing editing time by approximately an order of magnitude.
Qualitatively, it produces more consistent appearance across novel views,
reduced structural drift, and stable background preservation
on both DTU and RE10K.
These results demonstrate that enforcing cross-view consistency
in the image domain enables efficient and coherent sparse-view 3D editing
without per-scene optimization.
More results are provided in the supplementary material.

\subsection{Cross-view Editing consistency}
\label{sec:in2n_2d}

To assess the effectiveness of our cross-view fine-tuning
independently of the 3D lifting module,
we evaluate the fine-tuned FLUX backbone on IN2N
against prior text-guided editing methods.
This experiment isolates the editing component
and measures semantic alignment purely at the image level.

Following the standard IN2N protocol,
we apply the same editing prompts used in prior work.
We report 2D semantic alignment using
CLIP$_{sim}^{2D}$ and CLIP$_{dir}^{2D}$,
together with cross-view consistency measured
by DINO-based feature distance.

\begin{figure}[t!]
\centering
\includegraphics[width=\linewidth]{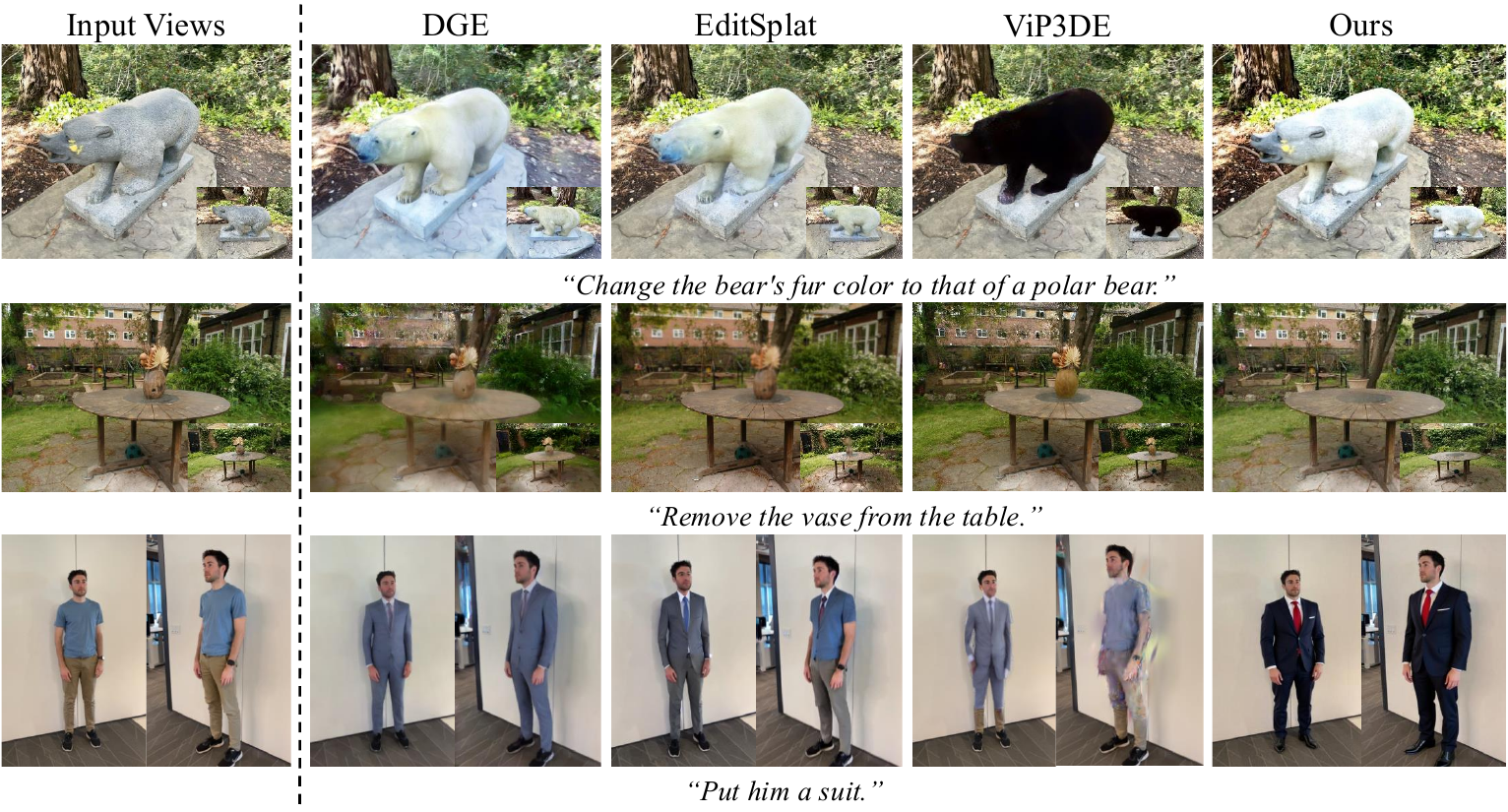}
\caption{\textbf{Cross-view 2D editing comparison on IN2N\cite{haque2023instruct}.}
Qualitative results comparing prior methods and our model.
Our approach produces cross-view edits that better align with the textual instruction.}
\label{fig:in2n}
\end{figure}

As shown in Tab.~\ref{tab:in2n_2d} and Fig.~\ref{fig:in2n},
our cross-view diffusion regularization preserves 2D editing fidelity
while consistently achieving stronger alignment with the target prompts.
These results indicate that the proposed consistency constraints
enhance editing robustness and stability,
rather than over-regularizing the generative process.

\begin{table}[t!]
\centering
\begin{tabular}{l c c c}
\toprule
Method 
& CLIP$_{sim}^{2D}$ $\uparrow$ 
& CLIP$_{dir}^{2D}$ $\uparrow$
& LEFL $\downarrow$ \\
\midrule
DGE\cite{chen2024dge} & 0.209 & 0.122 & 0.052 \\
EditSplat\cite{lee2025editsplat} & 0.207 & 0.125 & 0.048 \\
ViP3DE\cite{chen2025vip3dedit} & 0.193 & 0.096 & 0.069 \\
\midrule
Ours & \textbf{0.225} & \textbf{0.140} & \textbf{0.021} \\
\bottomrule
\end{tabular}
\caption{\textbf{Cross-view 2D editing performance on IN2N\cite{haque2023instruct}}.
We report both semantic alignment and cross-view consistency.
Our method improves text-edit alignment while
significantly reducing structural inconsistency across views.}
\label{tab:in2n_2d}
\vspace{-10pt}
\end{table}
\subsection{Ablation Studies}

We conduct ablation studies to analyze the contribution
of the proposed cross-view consistency regularization,
including Global Diffusion Feature Loss (GDFL),
Local Editing Feature Loss (LEFL),
and LoRA adaptation strategy.
All ablations are performed on top of the rectified-flow
fine-tuned editing backbone.

\paragraph{\textbf{Cross-View Regularization}}
\begin{table*}[t!]
\centering
\begin{tabular}{c c c c c c c}
\toprule
GDFL & LEFL 
& CLIP$_{sim}^{2D}$ $\uparrow$ 
& CLIP$_{dir}^{2D}$ $\uparrow$
& CLIP$_{sim}^{3D}$ $\uparrow$
& CLIP$_{dir}^{3D}$ $\uparrow$
& LEFL $\downarrow$ \\
\midrule
 &  & 0.118 & 0.186 & 0.197 & 0.051 & 0.078 \\
\checkmark &  & 0.127 & 0.201 & 0.213 & 0.074 & 0.046 \\
\checkmark & \checkmark & 0.140 & 0.225 & 0.231 & 0.086 & 0.021 \\
\bottomrule
\end{tabular}
\caption{\textbf{Ablation of cross-view regularization.}
We report both 2D and 3D semantic consistency together with
cross-view feature alignment.
GDFL improves global structural consistency across views,
while LEFL further enhances region-level coherence.
Importantly, consistency regularization does not degrade
2D semantic fidelity while significantly improving
3D semantic preservation and cross-view alignment.}
\label{tab:ablation_main}
\end{table*}

\begin{figure}[t]
\centering
\includegraphics[width=\linewidth]{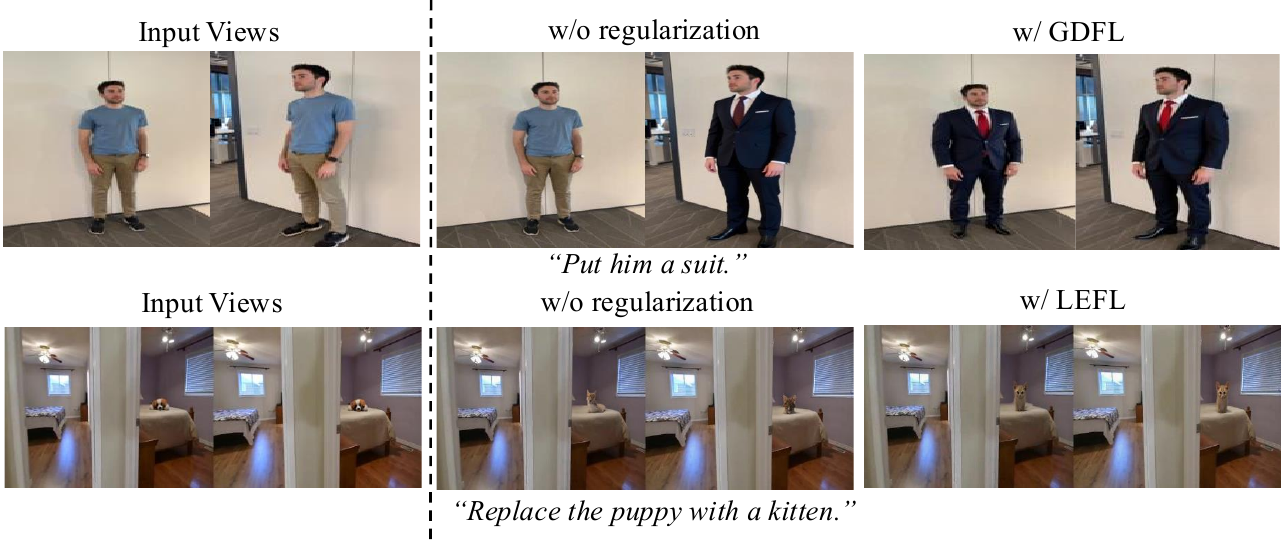}
\caption{\textbf{Qualitative comparison of cross-view regularization.}
Without cross-view regularization, edited views may exhibit inconsistent results:
either the edit appears in only one view, or both views are edited but with mismatched appearance.
Global Diffusion Feature Loss (GDFL) encourages global semantic alignment so that edits are consistently applied across views,
while Local Editing Feature Loss (LEFL) further enhances region-level coherence,
ensuring consistent appearance within the localized edited areas.}
\label{fig:consistency_ablation}
\end{figure}

Table~\ref{tab:ablation_main} and Fig.~\ref{fig:consistency_ablation} show the impact of progressively enabling the proposed cross-view consistency losses.
We report CLIP$_{sim}$, CLIP$_{dir}$, and cross-view feature distance to measure semantic and structural alignment.
GDFL significantly reduces global drift across viewpoints by aligning diffusion features,
while adding LEFL further improves region-level coherence without sacrificing semantic consistency.
Without regularization, edits exhibit view-dependent appearance and misalignment;
combining GDFL and LEFL produces globally consistent and locally coherent results under novel views.

\paragraph{\textbf{LoRA Configuration Analysis}}
\begin{figure}[t]
\centering
\includegraphics[width=\linewidth]{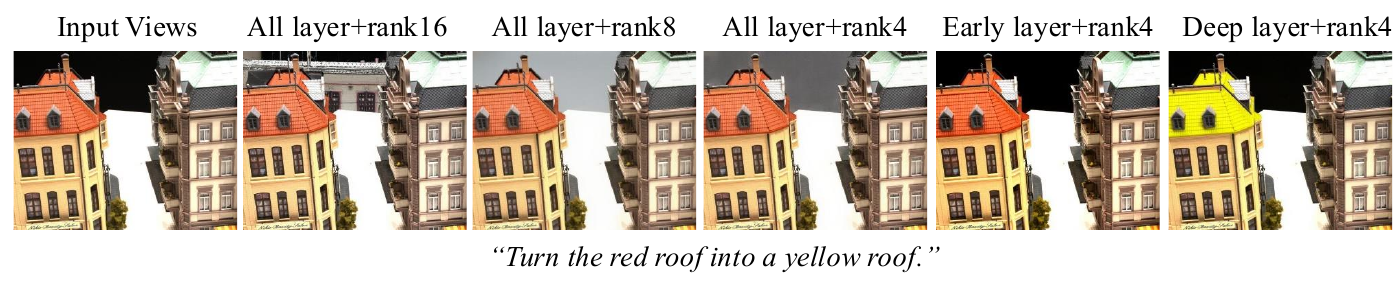}
\caption{\textbf{LoRA configuration comparison.}
We evaluate different rank adaptation strategies.
Applying high-rank LoRA globally induces semantic drift and weaker prompt alignment,
whereas low-rank adaptation restricted to deeper layers better preserves editing fidelity
and maintains cross-view structural consistency.}
\label{fig:lora_ablation}
\end{figure}

We compare several LoRA configurations, including global adaptation across all transformer layers with different ranks ($r=16, 8, 4$), early-layer adaptation ($r=4$), and deep-layer adaptation ($r=4$).
Qualitative results are shown in Fig.~\ref{fig:lora_ablation}.

Applying LoRA globally, especially with higher ranks, introduces noticeable semantic drift and weakens prompt adherence.
Reducing the rank mitigates this issue but still affects structural stability.
In contrast, restricting low-rank LoRA to deeper layers preserves the pretrained generative prior while enabling more precise, instruction-aligned edits.
The deep-layer, low-rank configuration produces the most accurate color transformation and the cleanest structural preservation across views.

Accordingly, we adopt the deep-layer LoRA with $r=4$, which achieves stronger editing fidelity and more stable cross-view consistency.


\subsection{Generalization}
Although the editing backbone is fine-tuned solely on RE10K,
\titlePrefix{} is directly applied to DTU and IN2N without any dataset-specific adaptation.
Despite substantial differences in scene scale and content,
our method maintains strong cross-view consistency on DTU (\cref{fig:dtu_re10k})
and achieves competitive qualitative performance on IN2N (\cref{fig:in2n}).
These results suggest that the proposed cross-view regularization
learns dataset-agnostic editing priors,
rather than overfitting to a particular scene distribution.


\section{Conclusion}
\label{sec:discussion}

We presented \titlePrefix{}, a feedforward sparse-view 3D editing framework
that enables text-guided manipulation from two input views.
Unlike prior methods that rely on iterative score distillation
or per-scene optimization, our approach enforces cross-view coherence
during training through feature regularization.

We introduce Global Diffusion Feature Loss (GDFL) for global alignment
and Local Editing Feature Loss (LEFL) for regional consistency.
Together, they enable single-step sparse-view editing
while preserving semantic fidelity and coherence.
By decoupling editing from scene refinement,
\titlePrefix{} generates an edited 3D Gaussian representation
in a single forward pass.

Future directions include extending the framework to multi-view editing
beyond two-view inputs, and integrating editing priors
into the reconstruction network to jointly adapt geometry
and appearance under textual guidance.

\newpage

%
%
\bibliographystyle{splncs04}
\bibliography{main}

\clearpage
\setcounter{page}{1}
\maketitlesupplementary

\section{Overview}
This supplementary material provides additional details and experimental results to complement the main paper.
First, we present implementation details of our method, including the design of the FLUX fine-tuning strategy, LoRA insertion positions, and training configurations. These details clarify how the editing backbone is adapted and how the proposed losses are implemented.

Next, we analyze representative failure cases to better understand the limitations of our approach. In particular, we highlight scenarios involving reflections and fine-grained texture transformations where cross-view consistency may degrade.

Finally, we present additional qualitative results across multiple datasets, including DTU, RE10K, and ScanNet. These examples demonstrate the robustness of our framework across diverse scenes and editing instructions, further illustrating its ability to produce coherent edits and maintain structural consistency across viewpoints.

\section{Implementation Details}
\subsection{FLUX Finetuning Design}
\justifying We adopt a parameter-efficient fine-tuning strategy based on LoRA for adapting the FLUX model.
Specifically, LoRA modules are inserted into the attention layers of selected transformer blocks within the diffusion backbone.

The LoRA layers are applied to the query-key-value projection (\textit{qkv}) and the output
projection (\textit{proj}) of the attention modules. This design enables efficient adaptation
of the attention mechanism while maintaining the majority of the pretrained model parameters frozen.

The insertion positions are summarized in Table~\ref{tab:lora_layers}. 
In particular, LoRA modules are applied to two DoubleStream blocks and four SingleStream blocks
within the transformer architecture.

\begin{table}[h!]
\centering
\begin{tabular}{ccc}
\hline
\textbf{Module Type} & \textbf{Layer Index} & \textbf{Insertion Position} \\
\hline
DoubleStream Block & 17, 18 & qkv + proj \\
SingleStream Block & 34, 35, 36, 37 & qkv + proj \\
\hline
\end{tabular}
\caption{LoRA insertion positions in the FLUX transformer architecture.}
\label{tab:lora_layers}
\end{table}

Our method leverages multiple feature representations for computing the training losses.
Specifically, diffusion features are used in the Global Diffusion Feature Loss (GDFL),
while DINO features are employed in the Local Editing Feature Loss (LEFL).
In addition, CLIP-based localization operates on patch-level image features.

The dimensionality of these features is summarized in Table~\ref{tab:feature_dims}.

\begin{table}[t!]
\centering
\begin{tabular}{cc}
\hline
\textbf{Feature Type} & \textbf{Dimension} \\
\hline
Diffusion Features (GDFL) & 1280 \\
DINO Features (LEFL) & 4096 \\
CLIP Localization Patch Size & $32 \times 32$ \\
\hline
\end{tabular}
\caption{Feature dimensions used in different components of the proposed method.}
\label{tab:feature_dims}
\vspace{-20pt}
\end{table}

\subsection{Training Details}

\justifying We train the model using the AdamW optimizer with a learning rate of $1\times10^{-4}$.
The optimizer uses $\beta_1=0.9$ and $\beta_2=0.999$, with $\epsilon=1\times10^{-8}$ and no weight decay.
The training is performed with a batch size of 1 for a total of 2000 optimization steps.

For the guidance losses, the Gaussian Diffusion Feature Loss (GDFL) operates on diffusion features extracted from the upsampling layer at index 1, with diffusion noise levels sampled from $t=261$ where $t \in [0,1000]$. 
The Localized Editing Feature Loss (LEFL) uses image patches of size $96 \times 96$ to enforce local editing consistency.

The overall loss is a weighted combination of the two objectives, where the weights of GDFL and LEFL are set to $0.4$ and $0.6$, respectively.

A summary is provided in Table.~\ref{tab:supp_training_config}.

\begin{table}[h!]
\centering
\small
\begin{tabular}{ll}
\toprule
\textbf{Category} & \textbf{Configuration} \\
\midrule

Optimizer & AdamW \\
Learning rate & $1\times10^{-4}$ \\
Betas $(\beta_1,\beta_2)$ & $(0.9, 0.999)$ \\
$\epsilon$ & $1\times10^{-8}$ \\
Weight decay & 0 \\

\midrule

Batch size & 1 \\
Training steps & 2000 \\

\midrule

Noise level for GDFL & $t = 261$ \\
Diffusion feature layer & Upsampling layer index 1 \\

\midrule

Patch size for LEFL & $96\times96$ \\

\midrule

GDFL weight & 0.4 \\
LEFL weight & 0.6 \\

\bottomrule
\end{tabular}
\caption{Training configuration and hyperparameters used in our method.}
\label{tab:supp_training_config}
\end{table}

\section{Failure Analysis}

\begin{figure}[htbp]
    \centering
    \includegraphics[width=1\linewidth]{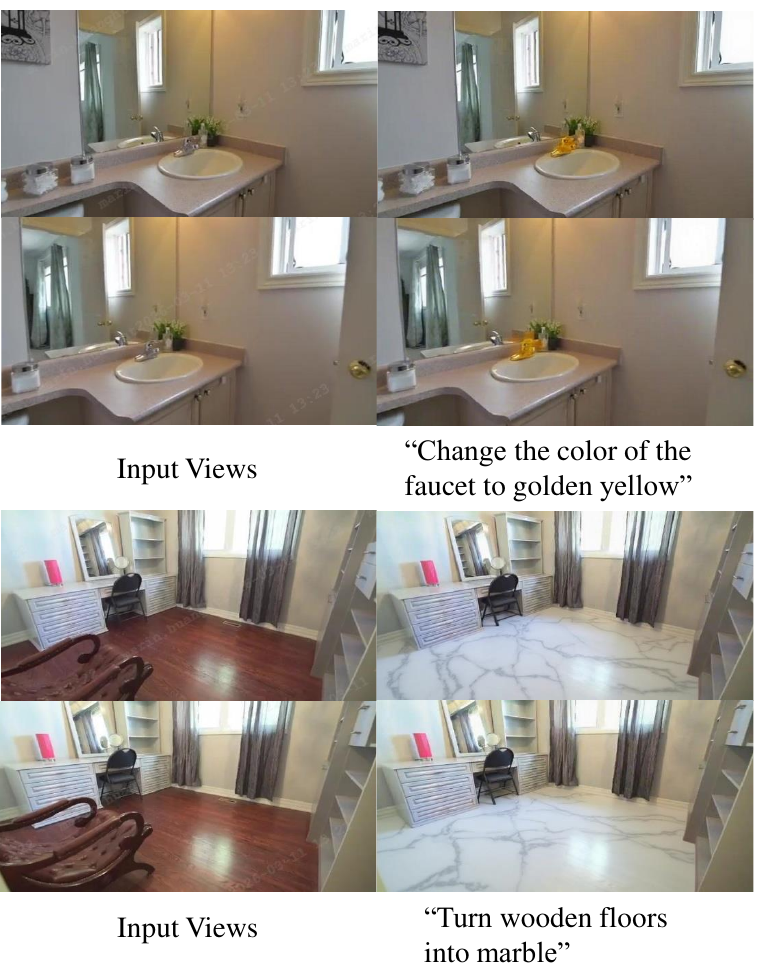}
\caption{\textbf{Failure case}
Our method occasionally produces inconsistent edits across views.
(Top) When editing the faucet color, the faucet visible in the mirror remains unedited, indicating that the model fails to propagate the edit to reflected regions.
(Bottom) When transforming wooden floors into marble, the generated marble texture exhibits inconsistency across views, leading to noticeable appearance variation under different viewpoints.
}
\label{fig:supp_failure_case}
\end{figure}

While our method generally produces consistent edits across views, certain challenging scenarios remain difficult. Fig.~\ref{fig:supp_failure_case} illustrates representative failure cases. 
In the top example, when editing the faucet color, the faucet visible in the mirror remains unchanged, indicating that the model does not always propagate edits to reflected regions. 
In the bottom example, when transforming wooden floors into marble, the generated marble texture shows noticeable inconsistency across views, leading to appearance variations under novel viewpoints. These cases highlight limitations in handling reflections and enforcing fine-grained texture consistency across views, suggesting directions for future improvements in cross-view reasoning and appearance modeling.


\section{More Qualitative Results}
\begin{figure}[!h]
    \centering
    \includegraphics[width=\linewidth]{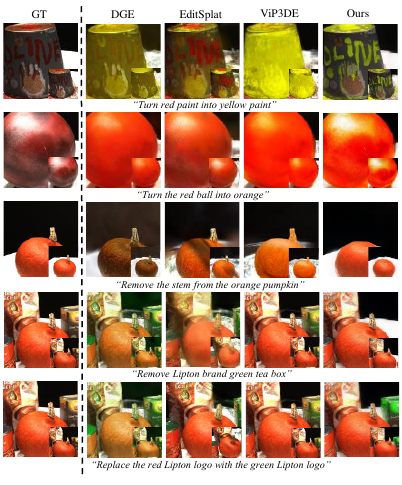}
    \caption{\textbf{Qualitative comparison on DTU~\cite{aanaes2016dtu}.}
We compare \titlePrefix{} with DGE~\cite{chen2024dge}, EditSplat~\cite{lee2025editsplat}, and ViP3DE\cite{chen2025vip3dedit}
under diverse text-guided editing instructions).
For each method, the large image and its inset correspond to \emph{two distinct novel views} of the edited 3D scene.
Our method produces more consistent color transformations,
better structural preservation, fewer cross-view artifacts, and more consistent object removal than prior approaches.
In contrast, prior methods exhibit color bleeding, or incomplete edits.
}
    \label{fig:supp_dtu}
\end{figure}

\begin{figure}[!h]
    \centering
    \includegraphics[width=\linewidth]{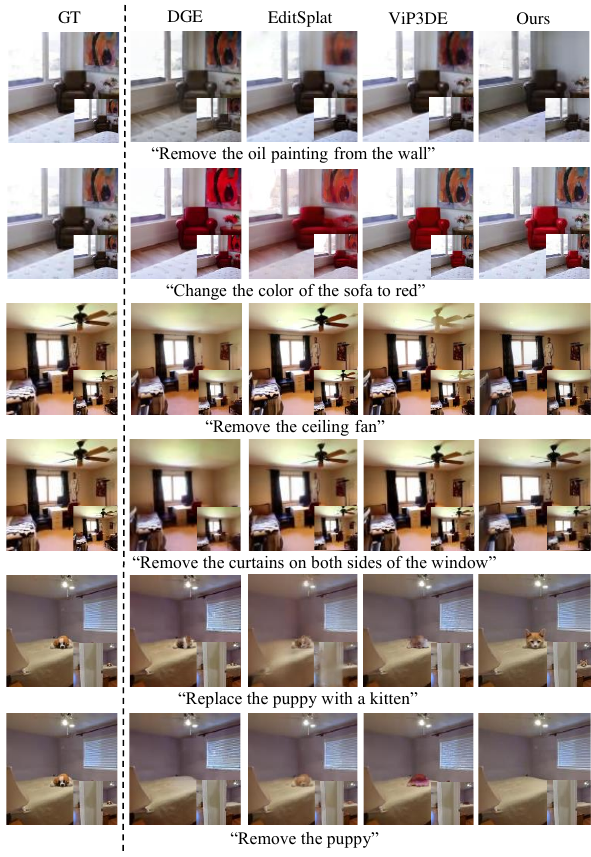}
\caption{\textbf{Qualitative comparison on RE10K~\cite{zhou2018stereo}.}
We compare \titlePrefix{} with DGE~\cite{chen2024dge}, 
EditSplat~\cite{lee2025editsplat}, and ViP3DE~\cite{chen2025vip3dedit}.
For each method, the large image and inset show two different novel views.
Our method achieves more consistent object removal and fewer cross-view artifacts than prior approaches.
}
    \label{fig:supp_re10k}
\end{figure}

\begin{figure*}[htbp]
    \centering
    \includegraphics[width=\textwidth,height=0.95\textheight,keepaspectratio]{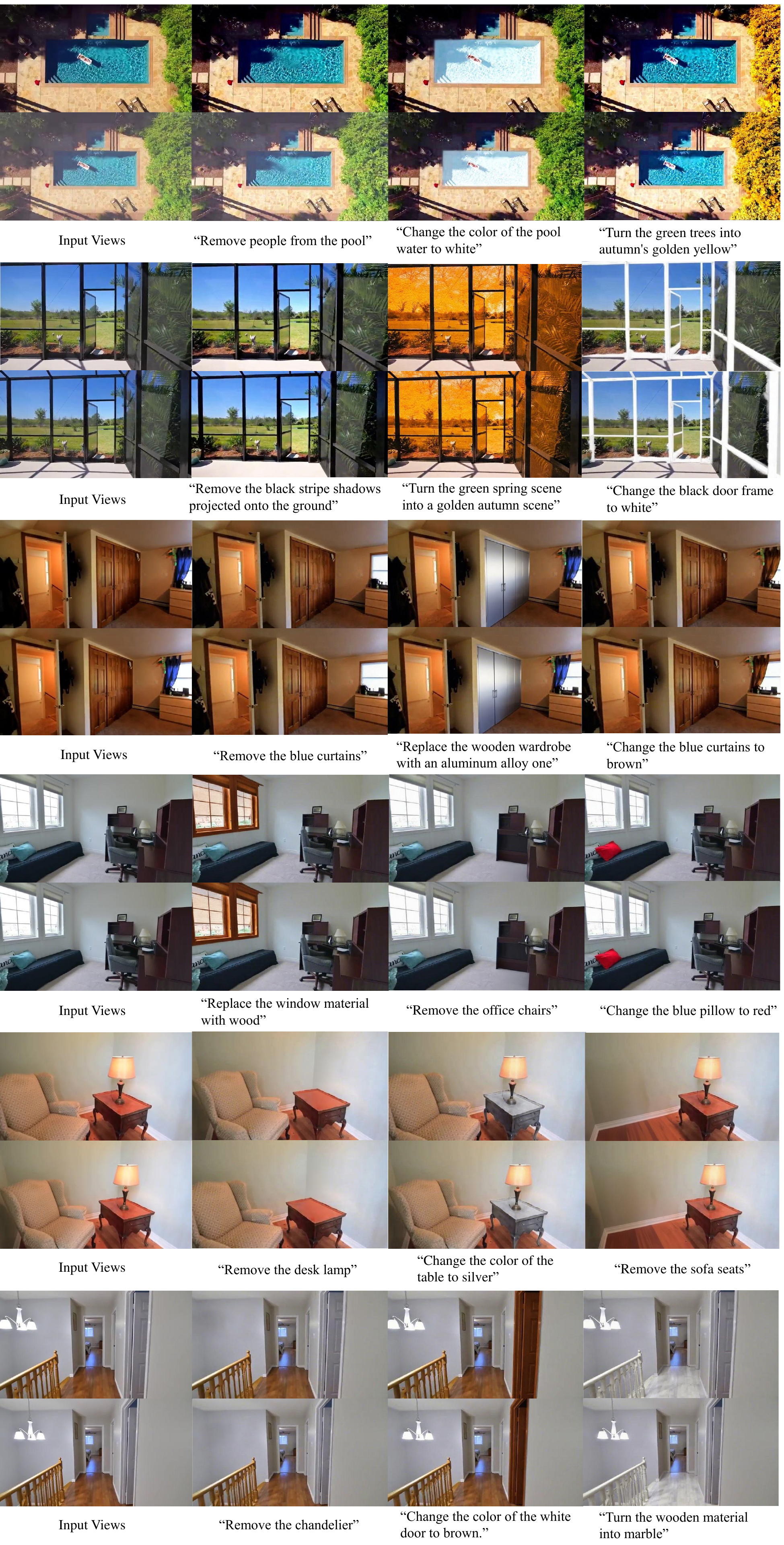}
\caption{\textbf{Additional editing results on RE10K~\cite{zhou2018stereo}.}}
\label{fig:supp_more_all}
\end{figure*}

\begin{figure*}[htbp]
    \centering
    \includegraphics[width=\textwidth,height=0.95\textheight,keepaspectratio]
    {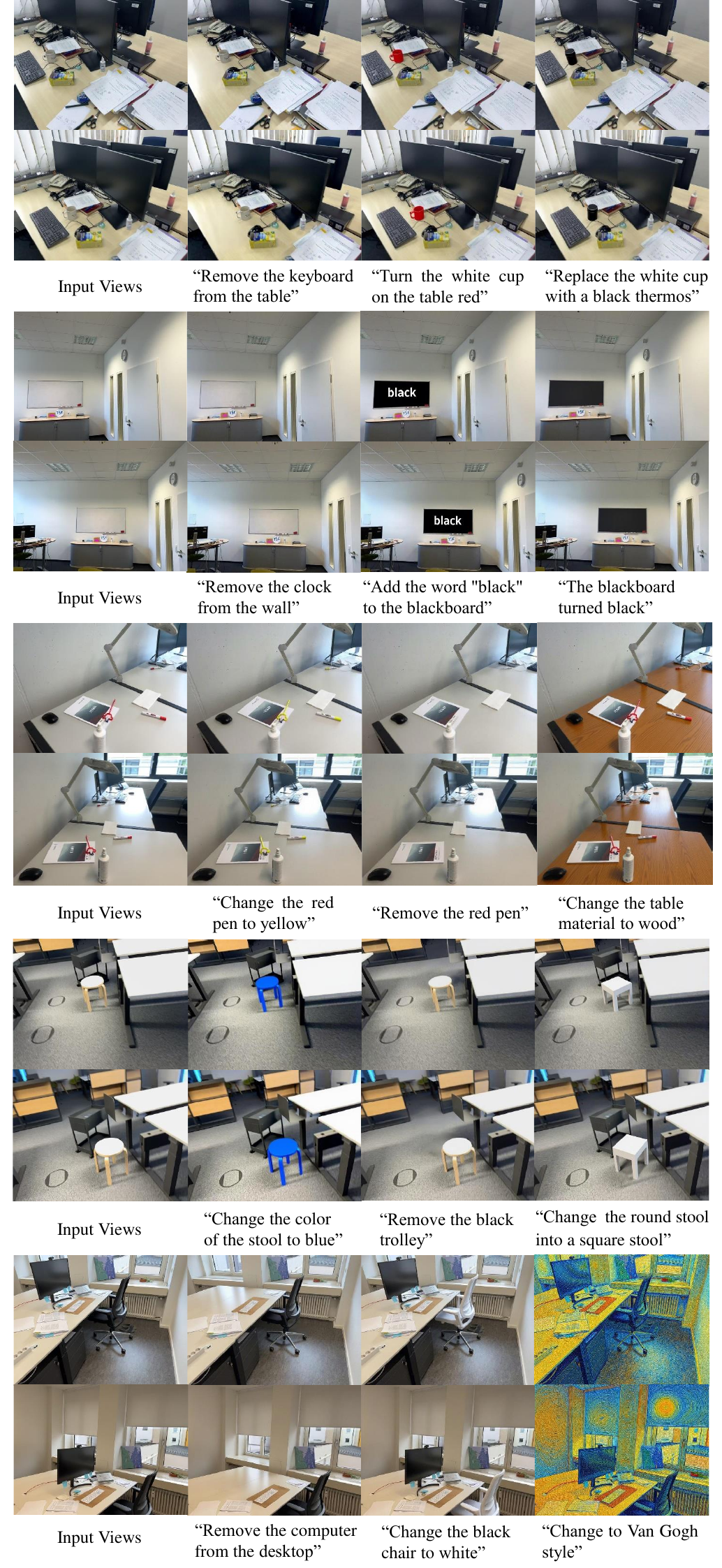}
\caption{\textbf{Additional editing results on ScanNet\cite{dai2017scannet}}
}
\label{fig:supp_scannet}
\end{figure*}

We present additional qualitative results to further demonstrate the robustness and visual quality of our method across diverse scenes and editing instructions. 
Fig.~\ref{fig:supp_dtu} and Fig.~\ref{fig:supp_re10k} provide extended comparisons with prior approaches on DTU\cite{aanaes2016dtu} and RE10K\cite{zhou2018stereo}, respectively, highlighting the cross-view consistency and structural preservation achieved by our method under novel viewpoints. 
Fig.~\ref{fig:supp_more_all} and Fig.~\ref{fig:supp_scannet} present additional editing results on RE10K and ScanNet\cite{dai2017scannet}, covering a wide range of editing scenarios. These examples further illustrate the ability of our framework to produce coherent edits across viewpoints while maintaining scene geometry and appearance consistency.

\end{document}